\documentclass[conference]{IEEEtran}
\IEEEoverridecommandlockouts

\usepackage{cite}
\usepackage{amsmath,amssymb,amsfonts}
\usepackage{graphicx}
\usepackage{textcomp}
\usepackage{xcolor}
\usepackage{tikz}
\usepackage{booktabs}
\usepackage{url}

\usetikzlibrary{arrows.meta,positioning,shapes.geometric}

\title{Transformer-Based Predictive Maintenance for Risk-Aware Instrument Calibration}

\author{
\IEEEauthorblockN{Adithya Parthasarathy}
\IEEEauthorblockA{\textit{California, USA} \\
 0009-0001-6839-9527}
 \and
\IEEEauthorblockN{ Aswathnarayan Muthukrishnan Kirubakaran}
\IEEEauthorblockA{\textit{California, USA} \\
 0009-0006-6652-2663}
 \and

\IEEEauthorblockN{Akshay Deshpande}
\IEEEauthorblockA{\textit{California, USA } \\
0009-0002-3007-3393}
\and
\IEEEauthorblockN{Ram Sekhar Bodala}
\IEEEauthorblockA{\textit{Delaware, USA} \\
0009-0005-4646-6679}

\and
\IEEEauthorblockN{Suhas Malempati}
\IEEEauthorblockA{\textit{South Carolina, USA} \\
0009-0009-3855-0423}

\and
\IEEEauthorblockN{Nachiappan Chockalingam	}
\IEEEauthorblockA{\textit{Massachusetts, USA} \\
 0009-0007-4275-3771} 
 \and

\IEEEauthorblockN{ Vinoth Punniyamoorthy}
\IEEEauthorblockA{\textit{Texas, USA} \\
 0009-0009-3719-4949}
  \and

\IEEEauthorblockN{ Seema Gangaiah Aarella}
\IEEEauthorblockA{\textit{Texas, USA} \\
 0009-0006-8807-3352}
}

\begin{document}
\maketitle

\begin{abstract}
Accurate calibration is essential for instruments whose measurements must remain traceable, reliable, and compliant over long operating periods. Fixed-interval programs are easy to administer, but they ignore that instruments drift at different rates under different conditions. This paper studies calibration scheduling as a predictive maintenance problem: given recent sensor histories, estimate time-to-drift (TTD) and intervene before a violation occurs. We adapt the NASA C-MAPSS benchmark into a calibration setting by selecting drift-sensitive sensors, defining virtual calibration thresholds, and inserting synthetic reset events that emulate repeated recalibration. We then compare classical regressors, recurrent and convolutional sequence models, and a compact Transformer for TTD prediction. The Transformer provides the strongest point forecasts on the primary FD001 split and remains competitive on the harder FD002--FD004 splits, while a quantile-based uncertainty model supports conservative scheduling when drift behavior is noisier. Under a violation-aware cost model, predictive scheduling lowers cost relative to reactive and fixed policies, and uncertainty-aware triggers sharply reduce violations when point forecasts are less reliable. The results show that condition-based calibration can be framed as a joint forecasting and decision problem, and that combining sequence models with risk-aware policies is a practical route toward smarter calibration planning.
\end{abstract}


\section{Introduction}
Precision instruments in laboratories, observatories, and healthcare systems do not lose calibration at a uniform rate. Environmental conditions, usage intensity, and component aging all affect how quickly measurements drift. Even so, many organizations still rely on fixed calibration intervals because they are easy to document and audit. The drawback is that a static interval treats every instrument as if it aged in the same way. In practice, that creates two avoidable outcomes: some instruments are recalibrated earlier than necessary, while others remain in service after drift has become operationally relevant \cite{pendrill2014optimised,beges2004calibration,ncsliRP1}.

For high-value or regulated assets, calibration is not only a quality activity but also a scheduling problem. Taking an instrument offline consumes technician time, interrupts experiments or production, and may idle dependent workflows. Missing a needed calibration has the opposite cost profile: the instrument stays available, but the organization absorbs measurement risk, compliance exposure, and potential rework. A useful data-driven policy therefore has to do more than fit a degradation curve; it has to convert that forecast into decisions that trade calibration effort against violation risk.

This paper addresses that problem in two stages. First, we estimate TTD from multivariate sensor windows. Second, we use those forecasts inside a scheduling rule that decides when to calibrate. Because public calibration datasets with repeated drift and reset cycles are scarce, we adapt the C-MAPSS turbofan benchmark into a calibration surrogate with virtual thresholds and synthetic recalibration events. We then compare linear models, tree ensembles, CNN/LSTM/TCN baselines, and a compact Transformer. The Transformer is used as the primary point forecaster, while a quantile LSTM provides the conservative uncertainty estimates consumed by the risk-aware scheduling rule.

The paper makes four contributions. First, it defines a full calibration-oriented adaptation of C-MAPSS, including drift-sensor selection, virtual calibration thresholds, synthetic reset events, and TTD labels. Second, it shows that a lightweight Transformer can outperform both classical baselines and other sequence models on the main calibration split. Third, it evaluates scheduling with violation-aware costs instead of relying on regression metrics alone. Finally, it demonstrates that lower-quantile triggers provide a practical safety mechanism when point forecasts degrade on harder operating-condition splits.

\section{Background and Related Work}
Calibration management sits between metrology and maintenance. In a failure-prediction setting, the main question is how long an asset can continue operating before it breaks down. In calibration, the question is slightly different and often more demanding: how long an instrument can remain in service before its readings are no longer trustworthy for the task at hand. The relevant threshold is therefore not catastrophic failure but loss of acceptable measurement performance. This distinction matters because a calibration action is a restorative reset applied before failure, and because the downstream cost is often tied to data validity, traceability, or compliance rather than to mechanical replacement alone.

Most calibration-interval research focuses on selecting or adjusting a single service interval from historical records, uncertainty budgets, or conformity requirements. That work provides the institutional logic behind interval setting: organizations need defensible policies that can be documented, repeated, and audited. What it usually does not provide is a rolling decision mechanism that updates priority instrument by instrument as fresh condition data arrive. In other words, interval-planning methods are useful for setting policy, but they do not by themselves solve the online scheduling problem addressed here.

The predictive-maintenance literature offers a closer operational analogy. There, the aim is to convert condition-monitoring signals into estimates of remaining useful life, fault probability, or intervention urgency. The connection to calibration is direct: both settings require models that map evolving multivariate measurements to a forward-looking maintenance decision. The difference is that calibration is driven by a performance threshold that may precede failure by a wide margin, and successful intervention resets the process rather than terminating it. That reset-and-repeat structure is common in practice but rare in public benchmark datasets, which is why a surrogate formulation is necessary.

Among available benchmarks, C-MAPSS remains one of the most widely used testbeds for temporal degradation modeling \cite{saxena2008cmapss,zhang2019rul}. Its appeal is not that it perfectly matches the calibration domain, but that it provides long multivariate trajectories under several operating-condition regimes. Those trajectories make it possible to study sensor monotonicity, degradation forecasting, and generalization across easier and harder condition mixes. By inserting virtual calibration thresholds and synthetic resets, the benchmark can be reinterpreted as a repeated drift-and-restoration problem rather than a single run-to-failure task. That adaptation preserves the temporal structure that makes the benchmark useful while aligning the labels with a calibration decision process.

The model families used in this paper serve different purposes. Linear regression provides a minimum-complexity reference point and clarifies how much signal is available without nonlinear modeling. Tree ensembles such as random forest, XGBoost, and LightGBM are strong baselines because they can capture nonlinear interactions among sensors without requiring heavy feature engineering. When the sensor window is flattened into a tabular representation, those models often provide a difficult baseline to beat. Their limitation is that temporal order is represented only implicitly through feature position, which means they do not model sequence structure as naturally as dedicated temporal architectures.

Recurrent and convolutional sequence models address that limitation in complementary ways. LSTMs \cite{hochreiter1997lstm} are designed to accumulate temporal evidence through gated recurrence, which makes them natural candidates when current drift depends on a long sensor history. One-dimensional CNNs extract local temporal motifs efficiently, while TCNs extend that idea with dilation and causal structure so that the receptive field grows without losing training stability \cite{bai2018tcn}. These models are useful not only because they are common baselines, but because they embody different hypotheses about the signal: recurrent memory emphasizes sequential accumulation, whereas temporal convolutions emphasize repeated local patterns.

Transformers offer a third view of the same problem. Instead of propagating state step by step or relying on fixed convolutional kernels, self-attention allows the model to compare measurements across the whole window and decide which positions are most informative for the forecast \cite{vaswani2017attention}. In degradation modeling, that can be valuable when the most predictive part of the signal is not the most recent measurement alone, but a broader temporal pattern or interaction among sensors. Prior work on RUL estimation and time-series forecasting suggests that Transformer-style models can be especially effective when long-range context matters and when the temporal dependencies are not well summarized by local convolutions or a single recurrent state \cite{mo2021remaining,wen2022transformers}.

For calibration scheduling, however, the model architecture is only half of the problem. A scheduler must convert a forecast into an action under asymmetric costs. A prediction that is unbiased on average may still be operationally poor if it is overconfident near the threshold-crossing region, where underestimation of risk is most expensive. This is why the paper separates point forecasting from conservative decision support. Quantile regression provides a natural bridge between prediction and action because it estimates different parts of the TTD distribution rather than only its center \cite{koenker1978quantile}. Recent neural approaches extend that idea to complex spatiotemporal settings and show that calibrated quantiles can serve as practical uncertainty summaries \cite{rodrigues2020beyond,tagasovska2019single}. In our setting, the lower quantile is especially relevant because it answers the operational question, ``how soon could this instrument become unsafe if the forecast is optimistic?''

This perspective also explains the choice to evaluate policies, not just regressors. Cost-aware predictive maintenance emphasizes that model selection should reflect the decision objective rather than only aggregate prediction error \cite{christodoulou2018costaware}. The same principle applies more strongly to calibration, where the imbalance between avoidable downtime and out-of-calibration operation is central to the problem definition. The literature therefore motivates the structure adopted here: a time-series forecasting task grounded in predictive-maintenance methods, combined with a scheduling layer that treats uncertainty and intervention cost as first-class concerns.

Another domain challenge is that drift is rarely observed in isolation. Instruments operate under changing loads, ambient conditions, and usage patterns, so the same sensor trend may imply different calibration urgency under different regimes. This is one reason multi-condition benchmarks such as FD002 and FD004 are important in the present study. They force the model to separate true drift progression from operating-context variation. A calibration scheduler that performs well only under a single stable regime would have limited practical value, because real fleets almost always contain instruments that age differently across sites, operators, and duty cycles.

This motivates comparing multiple model classes instead of treating one architecture as self-evidently best. In some deployments, interpretability, tuning simplicity, and latency may favor tree ensembles; in others, richer temporal models may justify their complexity by reducing missed drift events. The relevant question is therefore not only which model predicts TTD most accurately in aggregate, but which model preserves the decision-relevant structure of the signal near the calibration threshold. That framing helps connect the background literature to the empirical results that follow: the comparison is not merely algorithmic, but operational.

\section{Problem Formulation}
Consider an instrument $i$ emitting a multivariate sensor vector $\mathbf{x}_{i,t} \in \mathbb{R}^d$ at cycle $t$. Let $\mathcal{S}_i$ denote the subset of sensors selected as drift indicators for that instrument class or operating-condition split. For each selected sensor $s \in \mathcal{S}_i$, a virtual threshold $\tau_{i,s}$ marks the onset of unacceptable drift. Because some sensors drift upward and others downward, threshold crossing is defined relative to the estimated direction of degradation. A calibration action restores the process near baseline, after which a new drift cycle begins.

The forecasting task is defined on sliding windows. Given the recent history
\begin{equation}
\mathbf{X}_{i,t-w+1:t} = [\mathbf{x}_{i,t-w+1}, \ldots, \mathbf{x}_{i,t}] \in \mathbb{R}^{w \times d},
\end{equation}
the model predicts the time-to-drift (TTD) at cycle $t$, i.e., the number of cycles until the next threshold crossing. In the adapted dataset, a time index is considered out of calibration if any selected drift sensor has crossed its threshold. The ground-truth target is therefore
\begin{equation}
y_{i,t} = \min \{\Delta \ge 0 : \exists s \in \mathcal{S}_i \text{ such that } x_{i,t+\Delta,s} \text{ crosses } \tau_{i,s}\},
\end{equation}
with $y_{i,t}=0$ at or beyond a crossing and, when no later crossing exists in the current synthetic run, the label set to the remaining distance to the end of that run. The learning problem is to estimate a function $f_\theta$ such that $\hat{y}_{i,t} = f_\theta(\mathbf{X}_{i,t-w+1:t})$.

The forecast is only useful if it can drive a maintenance decision. We therefore evaluate policies under the cost
\begin{equation}
\text{cost} = c_{\text{cal}} \cdot N_{\text{cal}} + c_{\text{vio}} \cdot N_{\text{vio}},
\end{equation}
where $N_{\text{cal}}$ is the number of preventive calibrations and $N_{\text{vio}}$ is the number of threshold violations. At deployment, a policy maps the forecast to a binary decision
\begin{equation}
a_{i,t} =
\begin{cases}
1, & \text{if } s_{i,t} \le m,\\
0, & \text{otherwise,}
\end{cases}
\end{equation}
where $a_{i,t}=1$ denotes scheduling a calibration, $m$ is a safety margin, and $s_{i,t}$ is either the point forecast $\hat{y}_{i,t}$ or a conservative score such as the lower predicted quantile. If a planning window has finite service capacity $K$, instruments can be ranked by $s_{i,t}$ so that the smallest predicted TTD values are serviced first. This formulation distinguishes prediction quality from operational quality: a model with low average regression error can still be a poor scheduler if it systematically misses the threshold-crossing regime or becomes overconfident near the intervention boundary.

\subsection{Methodological Overview}
The proposed methodology follows the same sequence as the operational problem. We first adapt raw multivariate trajectories into a calibration dataset by identifying drift-sensitive sensors, defining sensor-specific thresholds, and inserting synthetic resets that mimic recalibration. We then construct fixed-length windows and train models to predict the TTD attached to the last cycle in each window. Finally, those forecasts are converted into decisions through point-based or quantile-based rules and evaluated with a violation-aware cost model.

This decomposition is intentional. It separates three sources of performance: whether the adaptation produces labels that resemble calibration behavior, whether the forecasting model captures the temporal structure of drift, and whether the decision rule converts forecasts into useful interventions. The sections that follow therefore describe the methodology in that same order: data adaptation, model estimation, and scheduling policy design.

\section{Data and Adaptation}
\subsection{Calibration-Oriented Adaptation of C-MAPSS}
C-MAPSS provides multivariate run-to-failure engine trajectories, but it does not contain calibration events. We reinterpret each engine run as an instrument drift trajectory and adapt each FD00$x$ subset independently. Within each split, drift-sensitive sensors are selected by Spearman correlation with operating cycle. On FD001, sensors 11, 4, and 12 are the most monotonic, with absolute correlations 0.615, 0.605, and 0.593, respectively. The same ranking procedure is repeated for FD002--FD004 so that each subset uses the sensors whose drift pattern is clearest under its own operating conditions.

\begin{figure}[t]
  \centering
  \includegraphics[width=\linewidth]{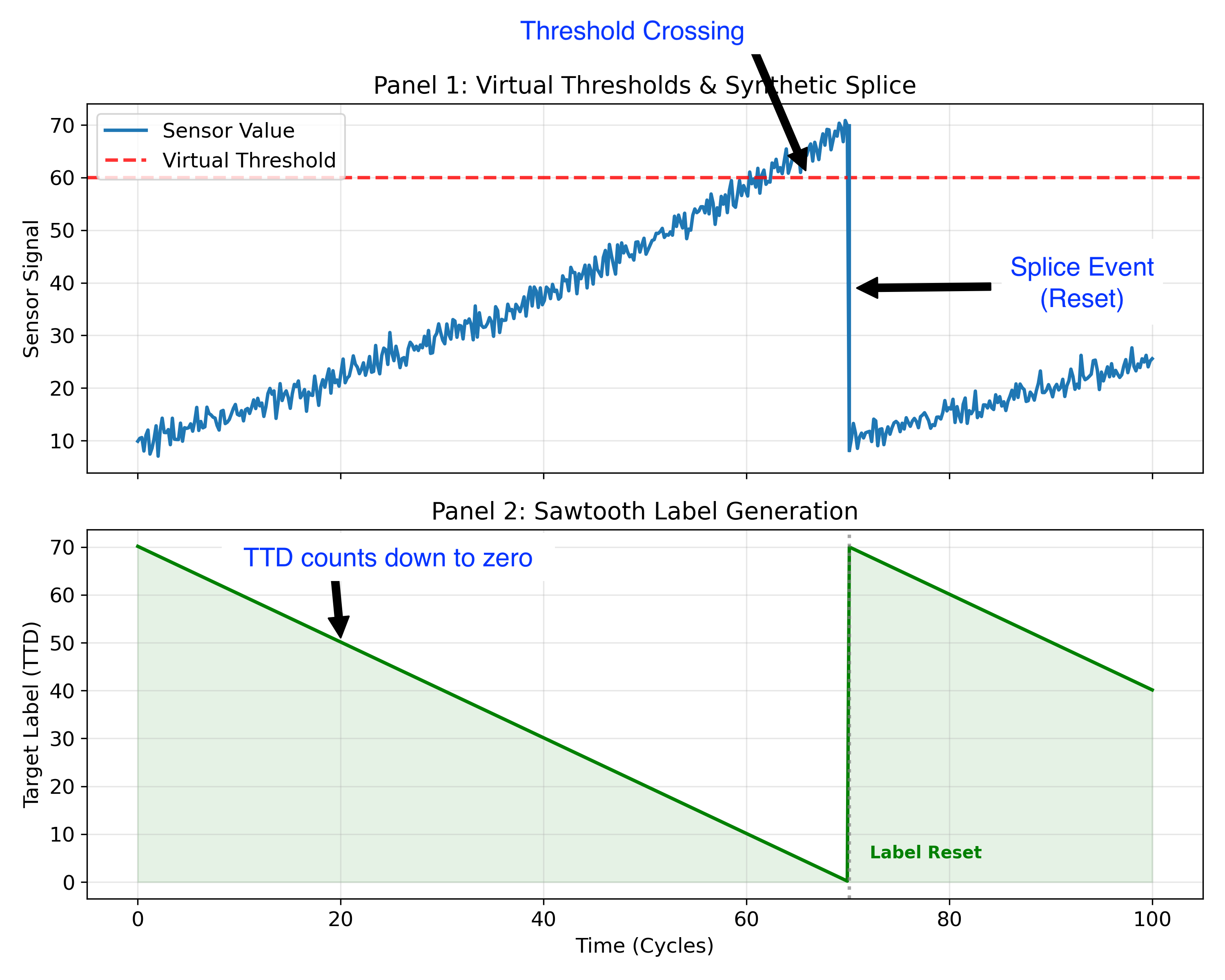}
  \caption{Calibration-oriented adaptation of C-MAPSS: sensor selection, virtual thresholds, synthetic resets, and TTD labeling.}
  \label{fig:data_adapt}
\end{figure}

Virtual calibration thresholds are then placed before end-of-life rather than at failure. For each selected sensor, the threshold is sampled uniformly from 55\%--80\% of the baseline-to-tail span. When a trajectory reaches the threshold, we synthesize a calibration event by resetting the selected drift sensors to baseline plus noise and allowing them to drift again. Some resets are generated by stitching in a perturbed early-life segment from another engine, which helps emulate imperfect recalibration and variable post-maintenance behavior. Up to three resets are inserted per run, producing repeated drift-reset cycles and multiple calibration opportunities from a single engine trace. Figure~\ref{fig:data_adapt} summarizes this transformation.

\subsection{Window Construction and Labels}
After adaptation, each trajectory is converted into overlapping 40-cycle windows with stride 1. Splits are performed at the engine level so that windows from the same engine do not leak across train and validation sets; 75\% of engines are used for training and 25\% for validation. Features are standardized with training statistics only. The regression target is TTD, defined as the number of cycles remaining until the next threshold crossing after the end of the window. For scheduling, negative predictions are clipped to zero so that overdue instruments are treated as immediate priorities rather than assigned artificial negative lead times.

\section{Models}
We compare four standard regressors linear regression, random forest, XGBoost, and LightGBM against four sequence models: LSTM, 1D CNN, TCN, and Transformer. The tree models provide strong tabular baselines after flattening each sensor window, while the neural models preserve temporal structure directly.

The LSTM also serves as the uncertainty model. It predicts the 0.1, 0.5, and 0.9 TTD quantiles with the pinball loss, producing the lower-quantile forecast used by the risk-aware scheduler. The CNN and TCN provide local-temporal baselines, with the TCN using two dilated causal convolution blocks. Our primary point predictor is a compact Transformer with sinusoidal positional encodings, two encoder layers, four attention heads, and model dimension $d_{\text{model}}=64$. It is trained with SmoothL1 loss and AdamW using 100 warmup steps followed by cosine decay. Xavier initialization is applied to linear layers and LSTM input weights, orthogonal initialization to recurrent weights, and Kaiming initialization to convolutional layers. The intent is to test whether a small attention model can extract enough temporal structure to improve calibration decisions without resorting to a large architecture.

\section{Scheduling and Cost Modeling}
\begin{figure}[t]
  \centering
  \begin{tikzpicture}[font=\footnotesize, node distance=8mm and 10mm, >=Latex, line width=0.4pt]
    \tikzset{
      block/.style={draw, rounded corners=2pt, fill=black!5, minimum width=2.8cm, minimum height=1.1cm, align=center, inner sep=3pt},
      decision/.style={draw, diamond, aspect=1.6, inner sep=2pt, align=center, fill=black!5},
      note/.style={draw, rounded corners=2pt, fill=white, align=left, inner sep=3pt, font=\scriptsize}
    }

    \node[block] (window) {Sensor window\\$\mathbf{X}_{t-w+1:t}$};
    \node[block, below=of window] (model) {Forecast model\\(point or quantile)};
    \node[block, below=of model] (score) {Decision score\\$\hat{y}_t$ or $q_{0.1}$};
    \node[decision, below=of score] (test) {Trigger if\\score $\le m$};
    \node[block, below=6mm of test, fill=green!12] (cal) {Schedule calibration};
    \node[block, below=6mm of cal, fill=blue!8] (hold) {Continue monitoring};
    \node[note, below=6mm of hold] (cap) {Capacity limit: at most $K$ instruments can be calibrated in a planning window.};

    \draw[->] (window) -- node[right] {$w \times d$ sensors} (model);
    \draw[->] (model) -- (score);
    \draw[->] (score) -- (test);
    \draw[->] (test) -- node[right] {yes} (cal);
    \draw[->, bend left=72] (test) to node[left] {no} (hold);
    \draw[->, dashed] (hold.north) -- (cal.south);
    \draw[->, dashed] (cap.north) -- (hold.south);
  \end{tikzpicture}
  \caption{Decision flow for the predictive calibration policy.}
  \label{fig:scheduler}
\end{figure}
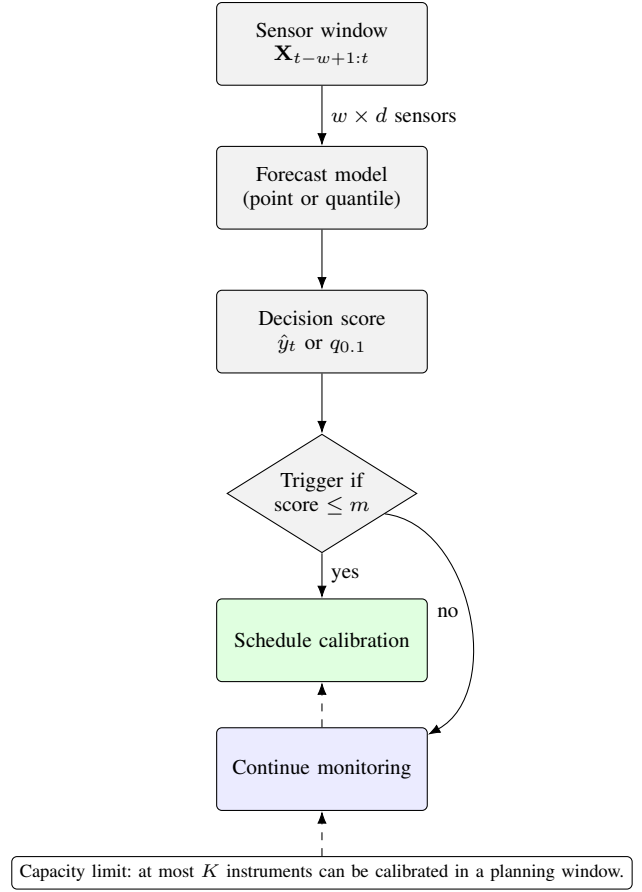

We evaluate four operational policies. The reactive policy calibrates only after a true violation. The fixed policy calibrates periodically. The predictive policy calibrates when the point forecast satisfies $\hat{y}_t \le m$, where $m$ is a lead-time margin. The uncertainty-aware policy replaces the point forecast with the lower quantile $q_{0.1}$, yielding a more conservative trigger on noisy splits. Figure~\ref{fig:scheduler} illustrates the decision flow.

Violations are counted only from ground-truth TTD values, not from forecast proxies. All experiments use $c_{\text{cal}}=1$ and $c_{\text{vio}}=5$, reflecting the common deployment asymmetry in which an unnecessary calibration is inconvenient but operating out of calibration is worse \cite{christodoulou2018costaware}. This setup lets the evaluation focus on the practical trade-off between downtime and risk rather than on regression accuracy alone.

\section{Training Setup}
All experiments were run on a Tesla T4 GPU with 14.7 GB of memory. The window length is 40 and the stride is 1 for all neural models. The Transformer is trained for up to 40 epochs with learning rate $3\times10^{-4}$ and patience 6. LSTM, CNN, and TCN models are trained for up to 30 epochs with learning rate $10^{-3}$. Tree ensembles use moderate estimator counts to keep the comparison practical rather than aggressively tuned. Unless noted otherwise, the same adaptation and scheduling pipeline is reused across FD001--FD004 so that differences in outcome mainly reflect data difficulty rather than per-split reconfiguration.

\section{Results}
\subsection{Prediction Accuracy on FD001}
\begin{figure}[t]
  \centering
  \includegraphics[width=\linewidth]{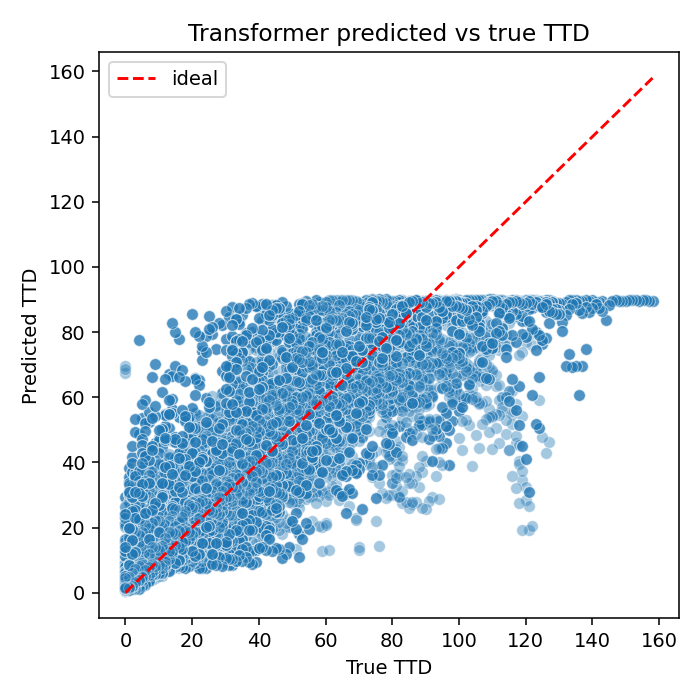}
  \caption{Predicted versus true TTD on FD001 for the Transformer.}
  \label{fig:pred}
\end{figure}

FD001 is the clearest calibration surrogate because it retains strong monotonic drift after adaptation. Table~\ref{tab:metrics} shows that the Transformer achieves the best validation accuracy, with MAE 13.84, RMSE 19.76, and $R^2=0.66$. LightGBM is the closest classical baseline at $R^2=0.64$, and the LSTM remains competitive at $R^2=0.60$. The CNN and TCN trail the other models on this split. Figure~\ref{fig:pred} provides a qualitative view of the Transformer predictions and shows that the forecast remains concentrated around the identity line.

\begin{table}[t]
\caption{Validation Metrics on C-MAPSS-Adapted FD001}
\label{tab:metrics}
\centering
\begin{tabular}{lccc}
\toprule
\textbf{Model} & \textbf{MAE} & \textbf{RMSE} & $\mathbf{R^2}$ \\
\midrule
Linear & 18.21 & 23.73 & 0.51 \\
Random Forest & 17.28 & 23.15 & 0.53 \\
XGBoost & 16.07 & 21.80 & 0.59 \\
LightGBM & 15.08 & 20.44 & 0.64 \\
LSTM & 15.78 & 21.39 & 0.60 \\
CNN & 19.57 & 25.66 & 0.43 \\
TCN & 19.75 & 25.97 & 0.41 \\
Transformer & \textbf{13.84} & \textbf{19.76} & \textbf{0.66} \\
\bottomrule
\end{tabular}
\end{table}

\subsection{Cross-Dataset Generalization}
The same architecture remains competitive on the other C-MAPSS subsets, although the ranking is split-dependent. On FD003 the Transformer again leads, reaching MAE 14.65 and $R^2=0.776$. On FD002 it matches the strongest tree baseline at roughly $R^2=0.29$, indicating that the benefit of attention narrows when drift monotonicity is weak. On FD004, LightGBM is the best model with $R^2=0.603$, while the Transformer reaches $R^2=0.486$. The broader pattern is still consistent: when drift retains a clear temporal structure, the Transformer is strongest; when operating conditions are more heterogeneous, tree models remain a credible alternative.

\subsection{Scheduling Outcomes}
The forecasting model is most useful when evaluated as a scheduler rather than as a regressor alone. On FD001, the point-based predictive policy reduces cost from 1734 under the reactive policy to 1193 while also lowering violations from 289 to 90. The quantile trigger is even safer, reducing violations to 26, but its conservative behavior raises the number of calibrations from 743 to 2386 and therefore increases total cost. This is the intended trade-off: lower quantiles protect against missed drift events at the expense of extra downtime.

\begin{table}[t]
\centering
\caption{Policy Cost Comparison on FD001}
\label{tab:policy}
\begin{tabular}{lccc}
\toprule
\textbf{Policy} & \textbf{Cal} & \textbf{Viol} & \textbf{Cost} \\
\midrule
Reactive & 289 & 289 & 1734 \\
Fixed & 417 & 289 & 1862 \\
Predictive & 743 & 90 & 1193 \\
Quantile@10 & 2386 & 26 & 2516 \\
\bottomrule
\end{tabular}
\end{table}

Table~\ref{tab:cross_policy} shows the same trade-off across FD002--FD004. FD003 benefits the most: both predictive policies nearly eliminate violations and dramatically improve cost relative to reactive or fixed scheduling. FD002 and FD004 are less stable, yet the uncertainty-aware rule still cuts violations sharply compared with the point policy. In other words, uncertainty becomes most valuable precisely on the splits where point accuracy is least reliable.

\begin{table}[t]
\caption{Cross-Split Policy Outcomes on C-MAPSS Datasets}
\label{tab:cross_policy}
\centering
\setlength{\tabcolsep}{4pt}
\begin{tabular}{l l c c c}
\toprule
\textbf{Data} & \textbf{Policy} & \textbf{Cal} & \textbf{Viol} & \textbf{Cost} \\
\midrule
FD002 & Reactive   & 3902  & 3902 & 23412 \\
      & Fixed      & 1043  & 3902 & 20553 \\
      & Predictive & 6063  & 2047 & 16298 \\
      & Quantile10 & 28341 & 152  & 29101 \\
\midrule
FD003 & Reactive   & 4805 & 4805 & 28830 \\
      & Fixed      & 503  & 4805 & 24528 \\
      & Predictive & 9075 & 13   & 9140 \\
      & Quantile10 & 9284 & 11   & 9339 \\
\midrule
FD004 & Reactive   & 3451 & 3451 & 20706 \\
      & Fixed      & 1317 & 3451 & 18572 \\
      & Predictive & 3965 & 2296 & 15445 \\
      & Quantile10 & 18021 & 823 & 22136 \\
\bottomrule
\end{tabular}
\end{table}

\section{Discussion and Future Work}
Two conclusions follow from the experiments. First, calibration scheduling should be treated as a joint forecasting-and-decision problem. The best regressor is not automatically the best operational policy unless its output is translated through a margin and a cost model. Second, a compact Transformer is a strong default when drift has clear temporal structure, but LightGBM remains a pragmatic alternative on heterogeneous splits. The results therefore support a practical model-selection view: choose the forecaster that best serves the scheduling objective, not just the lowest average regression error.

The most direct extension is to move the quantile head into the Transformer so that point and uncertainty estimates come from the same model. Additional gains are likely to come from per-condition normalization on FD002 and FD004, calibration of safety margins to site-specific cost ratios, and multi-seed evaluation to quantify variance. These are natural next steps within the present pipeline and would make the scheduler easier to tune for deployment.

\section{Conclusion}

This paper presented a calibration-oriented predictive maintenance framework that unifies time-to-drift (TTD) forecasting with risk-aware scheduling decisions, enabling a transition from static calibration intervals to dynamic, condition-based strategies. By adapting C-MAPSS into repeated drift-reset cycles, the study provides a realistic and reproducible benchmark for jointly evaluating prediction models and operational policies.

Empirical results demonstrate that the compact Transformer achieves the best predictive performance on FD001, attaining MAE of 13.84, RMSE of 19.76, and $R^2 = 0.66$, outperforming both classical models such as LightGBM ($R^2 = 0.64$) and sequence baselines. Its advantage persists on structured datasets such as FD003 ($R^2 = 0.776$), while remaining competitive on more heterogeneous splits (FD002 and FD004), where tree-based models provide strong alternatives. These findings confirm that attention-based models are particularly effective when drift exhibits clear temporal structure, while model selection remains data-dependent.

More importantly, evaluating models through a scheduling lens reveals substantial operational gains. On FD001, the predictive policy reduces total cost from 1734 (reactive) to 1193, while simultaneously lowering violations from 289 to 90, demonstrating the practical value of forecast-driven interventions. The uncertainty-aware (quantile) policy further reduces violations to 26, though at the expense of increased calibration frequency and higher cost, highlighting the trade-off between risk mitigation and operational efficiency. Across FD002--FD004, similar trends are observed: predictive scheduling consistently improves cost relative to fixed and reactive baselines, while quantile-based policies are most beneficial under high uncertainty and non-stationary conditions.

These results reinforce two key insights. First, calibration scheduling must be treated as a joint forecasting-and-decision problem, where success is defined by cost and risk outcomes rather than prediction error alone. Second, uncertainty modeling plays a critical role in safeguarding against missed drift events, particularly when model reliability varies across operating regimes.

In summary, the proposed framework demonstrates that Transformer-based forecasting combined with cost-aware and uncertainty-aware policies can significantly reduce unnecessary downtime while improving compliance assurance. 

\section*{Artifacts and Reproducibility}
Code, plots, tables, and summary artifacts are available in the public project repository \cite{riskawarecalibrationrepo}.

\end{document}